# MR-ULINS: A Tightly-Coupled UWB-LiDAR-Inertial Estimator with Multi-Epoch Outlier Rejection

Tisheng Zhang, Man Yuan, Linfu Wei, Yan Wang, Hailiang Tang*, Xiaoji Niu

*Abstract*—The LiDAR-inertial odometry (LIO) and the ultra-wideband (UWB) have been integrated together to achieve driftless positioning in global navigation satellite system (GNSS)-denied environments. However, the UWB may be affected by systematic range errors (such as the clock drift and the antenna phase center offset) and non-line-of-sight (NLOS) signals, resulting in reduced robustness. In this study, we propose a UWB-LiDAR-inertial estimator (MR-ULINS) that tightly integrates the UWB range, LiDAR frame-to-frame, and IMU measurements within the multi-state constraint Kalman filter (MSCKF) framework. The systematic range errors are precisely modeled to be estimated and compensated online. Besides, we propose a multi-epoch outlier rejection algorithm for UWB NLOS by utilizing the relative accuracy of the LIO. Specifically, the relative trajectory of the LIO is employed to verify the consistency of all range measurements within the sliding window. Extensive experiment results demonstrate that MR-ULINS achieves a positioning accuracy of around 0.1 m in complex indoor environments with severe NLOS interference. Ablation experiments show that the online estimation and multi-epoch outlier rejection can effectively improve the positioning accuracy. Besides, MR-ULINS maintains high accuracy and robustness in LiDAR-degenerated scenes and UWB-challenging conditions with spare base stations.

## I. INTRODUCTION

POSITIONING has played an increasingly important role in intelligent robots and autonomous vehicles in recent years. Inertial measurement unit (IMU) can measure the acceleration and angular velocity of the ego-motion. Hence, the inertial navigation system (INS) can be adopted by utilizing the IMU measurements to obtain high-frequency poses. Light detection and ranging (LiDAR) can be integrated with the IMU to construct the LiDAR-inertial odometry (LIO), and thus the accumulative errors of IMU can be reduced. However, LIOs are still dead-reckoning (DR) systems by using the frame-to-frame (F2F) data associations [1], and thus the position may drift over time. The drift can be eliminated by matching with a prebuilt high-definition map, which is of high expense and difficult to obtain. Besides, the prebuilt map may also fail in frequently changing environments, such as warehouses. Another method to eliminate cumulative errors is to integrate with absolute positioning sensors such as the global navigation satellite system (GNSS) [2] and ultra-wideband (UWB) [3]. The GNSS can achieve centimeter-level absolute positioning in open-sky environments, while it may fail in indoor environments. In contrast, UWB can provide drift-free indoor positioning by measuring distances from the UWB tag to base stations.

The UWB and IMU have been integrated in many studies. Zheng *et al.* [4] tightly integrated the UWB and IMU based on a graph optimization model. Wang *et al.* [5] fused the UWB and IMU with a particle filter. In [6] and [7], the UWB and IMU were combined with the unscented Kalman filter. However, the low DR accuracy of INS is insufficient for practical applications. In [8], UWB ranges were used to reduce the visual odometry (VO) drift with a pose-graph optimization scheme. VIR-SLAM [9] proposed a double-layer sliding-window algorithm to combine visual-inertial odometry (VIO) and UWB. There are also many studies that integrate LiDAR with UWB, due to the rapid development of low-cost solid-state LiDARs. Feng *et al.* [10] utilized the difference between UWB and LiDAR-based distances to estimate the UWB positioning error and then correct the positioning. Zhou *et al.* [11] proposed a LiDAR-UWB fusion algorithm by minimizing the sum of the Mahalanobis norm of all measurement residuals. Wang *et al.* [12] presented a UWB-LiDAR tightly-coupled positioning method using an optimized particle filter. Liu *et al.* [13] employed an iterative error state Kalman filter (IESKF) to tightly integrate UWB, LiDAR, and IMU raw measurements. LIRO [14] tightly fused the LiDAR, IMU, and UWB within an optimization framework. Hu *et al.* [15] designed a cooperative optimization module to cope with the cumulative error in LiDAR odometry (LO) through UWB observations.

However, the above UWB-LiDAR(-inertial) integrated methods ignore UWB systematic range errors, which may decrease positioning accuracy. The systematic range errors were pre-calibrated in [6], [9], and [16]. However, pre-calibration inevitably increases the costs, and the range errors may vary with environmental factors such as temperature and humidity. Therefore, it is necessary to study the modeling and online estimation of UWB systematic range errors. Besides, UWB non-line-of-sight (NLOS) signals seriously affect the accuracy of range measurements, but it is simply dealt with or even not considered in the above studies.

There are studies on UWB NLOS for UWB-only systems.

This research is partly funded by the Major Program (JD) of Hubei Province (2023BAA026) and the National Natural Science Foundation of China (No.42374034). (*Corresponding author: Hailiang Tang.*)

Tisheng Zhang and Xiaoji Niu are with the GNSS Research Center, Wuhan University, Wuhan 430079, China, and also with the Hubei Luojia Laboratory, Wuhan 430079, China (e-mail: zts@whu.edu.cn; xjniu@whu.edu.cn).

Man Yuan, Linfu Wei, Yan Wang, and Hailiang Tang are with the GNSS Research Center, Wuhan University, Wuhan 430079, China (e-mail: yuanman@whu.edu.cn; weilf@whu.edu.cn; wystephen@whu.edu.cn; thl@whu.edu.cn).

Guvenc *et al.* [17] identified the NLOS signals based on the multipath channel statistics. Marano *et al.* [18] used a support vector machine (SVM) classifier to distinguish between LOS and NLOS signals. Jiang *et al.* [19] employed the deep learning method CNN-LSTM in the NLOS/LOS signal classification. In multi-sensor fusion systems, UWB NLOS can be mitigated by fully utilizing the observation from other sensors. In [4], NLOS was mitigated by a residual test, which utilized the short-term high-accuracy property of INS. Considering that NLOS would only produce outliers with higher values than the true distance, VIR-SLAM [9] applied an acceptable bound based on the maximum motion to reduce NLOS measurements. In [16], the IGG3 equivalent weight function of the range residuals was used to exclude or reweight range measurements with potential NLOS errors. Liu *et al.* [13] calculated the difference of the UWB range and the distance from the tag to the base station using the estimated pose of LIO, and conducted a chi-square test and a residual test to mitigate NLOS. Wang *et al.* [12] incorporated LiDAR measurements by an optimized particle filter and offered the distance estimation between the positioning system and the UWB base station, thus mitigating NLOS.

Nevertheless, the above methods rely much on the absolute pose. If the estimated pose is not accurate, their performance will decrease significantly. Especially, under conditions with sparse UWB base stations, the accurate absolute pose cannot be obtained initially and the above methods become ineffective. Therefore, it is necessary to study more effective and robust algorithms, such as the random sample consensus (RANSAC) [20], to mitigate the impact of UWB NLOS for multi-sensor fusion.

Motivated by the above issues, a tightly-coupled UWB-LiDAR-inertial estimator with multi-epoch outlier rejection is proposed in this paper, named MR-ULINS. A F2F LiDAR data-association method is employed to construct a consistent LIO [1]. Hence, the absolute positioning sensor UWB can be seamlessly incorporated. The multi-state constraint Kalman filter (MSCKF) [21] is employed for state estimation due to its higher efficiency and comparable accuracy than graph optimization, while it can also contribute to UWB outlier rejection. The main contributions of our work are as follows:
- We present a UWB-LiDAR-inertial estimator within the MSCKF framework that tightly integrates the IMU, LiDAR, and UWB range measurements. The UWB systematic range errors are modeled as the bias and scale factor, and they are estimated online to further improve the positioning accuracy.
- We design a RANSAC-based multi-epoch outlier-rejection algorithm for UWB ranges. The short-term high-precision relative trajectory of LIO is utilized to verify the consistency of multi-epoch UWB ranges, thereby mitigating UWB NLOS. The proposed algorithm is immune to the absolute pose error and is effective in conditions with sparse UWB base stations.
- We conduct extensive real-world experiments in LOS, NLOS, and sparse base-station environments. The results demonstrate that the proposed MR-ULINS yields high

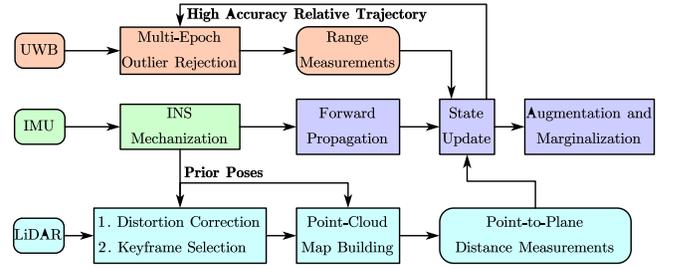

Fig. 1. System overview of the proposed MR-ULINS.

accuracy and effectively mitigates the impact of UWB systematic range errors and NLOS.

## II. SYSTEM OVERVIEW

The system overview of the proposed MR-ULINS is shown in Fig. 1. We adopt an INS-centric processing pipeline. The system runs in a local world frame, whose origin is the initial position of the IMU, and the z-axis is chosen to be gravity-aligned. Once the INS is initialized, the INS mechanization is performed to obtain the high-frequency INS poses, and the state vector and its covariance are forward propagated.

When a LiDAR frame is received, the point-cloud distortion is corrected with prior INS poses. Besides, the INS poses are employed for keyframe selection. If the relative motion or time interval between the current frame and the previous keyframe exceeds the set threshold, the current frame is considered as a keyframe. When a LiDAR keyframe is selected, all non-keyframes between the previous keyframe and the current keyframe are merged into the previous keyframe to construct the keyframe point-cloud map. Then, the frame-to-frame association is conducted between the current keyframe and historical keyframe point-cloud maps within the LiDAR sliding window. Thus, the LiDAR F2F point-to-plane distance measurement model can be utilized to update the MSCKF state.

Meanwhile, the UWB range measurements are compensated by the estimated systematic range errors. The corrected range measurements and current pose are used to construct a UWB keyframe. When the number of range measurements in the UWB sliding window is sufficient for UWB update, the proposed multi-epoch outlier rejection is conducted to remove the outliers caused by errors such as NLOS. Finally, the refined range measurements are employed to update the MSCKF state and eliminate the drift of the LIO.

After the LiDAR/ UWB measurement update, the IMU pose state at the corresponding time is augmented into the MSCKF state vector. Besides, the marginalization is conducted when the LiDAR/UWB sliding window exceeds its maximum length to maintain computational efficiency.

## III. MSCKF-BASED UWB-LIDAR-INERTIAL ESTIMATOR

### A. State Vector

The MSCKF state vector $\delta x$ consists of error states from INS, LiDAR-IMU extrinsic parameters, UWB systematic range errors, LiDAR keyframes, and UWB keyframes, which are represented by $\delta \boldsymbol{x}_I$, $\delta \boldsymbol{x}_I^b$, $\delta \boldsymbol{x}_{UWB}$, $\delta \boldsymbol{x}_{KF_{LiDAR}}$ and $\delta \boldsymbol{x}_{KF_{UWB}}$, respectively. The specific forms can be defined as:

$$\delta \boldsymbol{x} = [\delta \boldsymbol{x}_I, \delta \boldsymbol{x}_I^b, \delta \boldsymbol{x}_{UWB}, \delta \boldsymbol{x}_{KF_{LiDAR}}, \delta \boldsymbol{x}_{KF_{UWB}}]^T, \quad (1)$$

where

$$\begin{aligned}
\delta \boldsymbol{x}_I &= [\delta \boldsymbol{\theta}_b^w, \delta \boldsymbol{p}_b^w, \delta \boldsymbol{v}^w, \delta \boldsymbol{b}_g, \delta \boldsymbol{b}_a], \\
\delta \boldsymbol{x}_I^b &= [\delta \boldsymbol{\theta}_I^b, \delta \boldsymbol{p}_I^b], \\
\delta \boldsymbol{x}_{UWB} &= [\delta s_{u_1}, \delta s_{u_2}, \cdots, \delta s_{u_q}, \delta b_{u_1}, \delta b_{u_2}, \cdots, \delta b_{u_q}], \quad (2)\\
\delta \boldsymbol{x}_{KF_{LiDAR}} &= [\delta \boldsymbol{x}_{KF_{LiDAR},0}, \delta \boldsymbol{x}_{KF_{LiDAR},1}, \cdots, \delta \boldsymbol{x}_{KF_{LiDAR},N-1}], \\
\delta \boldsymbol{x}_{KF_{UWB}} &= [\delta \boldsymbol{x}_{KF_{UWB},0}, \delta \boldsymbol{x}_{KF_{UWB},1}, \cdots, \delta \boldsymbol{x}_{KF_{UWB},M-1}].
\end{aligned}$$

Here, $w$, $b$, $l$, and $u$ represent the world frame, IMU frame, LiDAR frame, and UWB tag frame, respectively; $q$ is the number of UWB base stations; $N$ and $M$ are the lengths of the LiDAR sliding window and UWB sliding window, respectively; $\delta \boldsymbol{\theta}_b^w$, $\delta \boldsymbol{p}_b^w$ and $\delta \boldsymbol{v}^w$ are the errors of current IMU attitude, position, and velocity, respectively; $\delta \boldsymbol{b}_g$ and $\delta \boldsymbol{b}_a$ are the bias errors of the gyroscope and accelerometer, respectively; $\delta \boldsymbol{\theta}_I^b$ and $\delta \boldsymbol{p}_I^b$ are the errors of LiDAR-IMU extrinsic parameters; $\delta s_{u_k}$ and $\delta b_{u_k}$ are the scale factor error and bias error of the $k$-th UWB base station; $\delta \boldsymbol{x}_{KF_{LiDAR},k}$ and $\delta \boldsymbol{x}_{KF_{UWB},k}$ can be expressed as

$$\delta \boldsymbol{x}_{KF_{LiDAR},k} = [\delta \boldsymbol{\theta}_{l_k}^w, \delta \boldsymbol{p}_{l_k}^w], \quad (3)$$

$$\delta \boldsymbol{x}_{KF_{UWB},k} = [\delta \boldsymbol{\theta}_{u_k}^w, \delta \boldsymbol{p}_{u_k}^w], \quad (4)$$

where $\delta \boldsymbol{\theta}_{l_k}^w$ and $\delta \boldsymbol{p}_{l_k}^w$ are the attitude and position errors of the $k$-th LiDAR keyframe; $\delta \boldsymbol{\theta}_{u_k}^w$ and $\delta \boldsymbol{p}_{u_k}^w$ are the attitude and position errors of the $k$-th UWB keyframe.

The true state $\boldsymbol{x}$ can be obtained using the estimated state $\hat{\boldsymbol{x}}$ and the error state $\delta \boldsymbol{x}$ as

$$\boldsymbol{x} = \hat{\boldsymbol{x}} \boxplus \delta \boldsymbol{x}. \quad (5)$$

For the attitude error $\delta \boldsymbol{\theta}$, the operator $\boxplus$ is given by

$$\mathbf{R} = \hat{\mathbf{R}} \mathrm{Exp}(\delta \boldsymbol{\theta}) \approx \hat{\mathbf{R}}(\mathbf{I} + (\delta \boldsymbol{\theta}) \times), \quad (6)$$

where $\mathbf{R}$ and $\hat{\mathbf{R}}$ denote the true and estimated rotation matrix, respectively; $\mathrm{Exp}$ is the exponential map [22]; $(\cdot) \times$ denotes the skew-symmetric matrix of the vector belonging to $\mathbb{R}^3$ [23]. For other states, the operator $\boxplus$ is equivalent to Euclidean addition, i.e., $\boldsymbol{a} = \hat{\boldsymbol{a}} + \delta \boldsymbol{a}$.

When the IMU measurement is available, the INS mechanization [24] is conducted to output the high-frequency prior pose. Meanwhile, the forward propagation of the whole error state and its covariance is performed using the standard error-state Kalman filter (ESKF) prediction formula [25].

*B. LiDAR Measurement Model*

The LiDAR frame-to-frame (F2F) data association is conducted between the current LiDAR keyframe and the historical keyframe point-cloud maps to construct relative pose constraints [1], as shown in Fig. 2. The specific steps for the F2F data association are as follows. For a point $\boldsymbol{p}^{l_N}$ in the current LiDAR keyframe, it is firstly projected onto the $k$-th

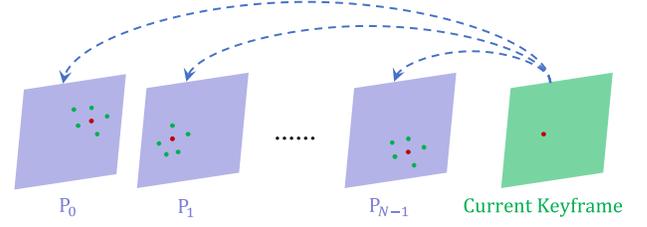

$P_k$ The LiDAR keyframe point-cloud map
• The points in the current keyframe and its projection
• The nearest neighboring points in $P_k$.

Fig. 2. Illustration of the LiDAR frame-to-frame data association.

keyframe point-cloud map $P_k$ with the current keyframe pose $\{\hat{\mathbf{R}}_{l_N}^w, \hat{\boldsymbol{p}}_{l_N}^w\}$ and the $k$-th keyframe pose $\{\hat{\mathbf{R}}_{l_k}^w, \hat{\boldsymbol{p}}_{l_k}^w\}$. The projection point $\boldsymbol{p}^{l_k}$ can be expressed as

$$\begin{aligned}
\boldsymbol{p}^{l_k} &= \hat{\mathbf{R}}_{l_N}^{l_k} \boldsymbol{p}^{l_N} + \hat{\boldsymbol{p}}_{l_N}^{l_k}, \\
\hat{\mathbf{R}}_{l_N}^{l_k} &= (\hat{\mathbf{R}}_{l_k}^w)^T \hat{\mathbf{R}}_{l_N}^w, \quad (7)\\
\hat{\boldsymbol{p}}_{l_N}^{l_k} &= (\hat{\mathbf{R}}_{l_k}^w)^T (\hat{\boldsymbol{p}}_{l_N}^w - \hat{\boldsymbol{p}}_{l_k}^w).
\end{aligned}$$

Then, five nearest neighboring points of $\boldsymbol{p}^{l_k}$ are searched in $P_k$ by a KD-tree [26], and denoted as $\{\boldsymbol{p}_i^{l_k} \mid i \in [1,5]\}$. The five points are used to fit a plane, which can be expressed as

$$\boldsymbol{n}^T \boldsymbol{p} + d = 0, \quad (8)$$

where $\boldsymbol{p}$ is a point on the plane; $\boldsymbol{n}$ is the normalized normal vector of the plane; $d$ is the distance that satisfies Equation (14). Only when all the distances from points $\boldsymbol{p}^{l_k}$ and $\boldsymbol{p}_i^{l_k}$ to the plane are less than 0.1 m [1], $\boldsymbol{p}^{l_k}$ will be used to construct the LiDAR measurement model.

The LiDAR F2F measurement is the distance from $\boldsymbol{p}^{l_k}$ to the fitted plane $(\boldsymbol{n}, d)$. Since $\boldsymbol{p}^{l_k}$ is on the plane, the residual is equivalent to the point-to-plane distance and can be written as

$$r_{N,k}^{LiDAR}(\boldsymbol{p}^{l_N}, \delta \boldsymbol{x}) = \boldsymbol{n}^T \boldsymbol{p}^{l_k} + d. \quad (9)$$

The residual is a function of the current IMU pose errors $\{\delta \boldsymbol{\theta}_b^w, \delta \boldsymbol{p}_b^w\}$, LiDAR-IMU extrinsic errors $\{\delta \boldsymbol{\theta}_I^b, \delta \boldsymbol{p}_I^b\}$, and $k$-th LiDAR keyframe pose errors $\{\delta \boldsymbol{\theta}_{l_k}^w, \delta \boldsymbol{p}_{l_k}^w\}$. Therefore, we can derive the corresponding analytical Jacobians using the error-perturbation method [24]. With the residuals and the analytical Jacobians, the MSCKF error state and covariance can be updated with the standard ESKF update formula [25].

*C. UWB Measurement Model*

The UWB range measurements are obtained by measuring the time of arrival (TOA) of the signal [3], and a double-side two-way ranging (DS-TWR) method [27] is employed to eliminate the need for synchronization between base stations. The systematic range errors of the DS-TWR method mainly come from the antenna phase center offsets and the minimum clock drift between the tag and base station, and they need to be considered for precise positioning. Offline calibration methods cannot timely estimate the systematic range errors as they may be affected by environmental factors such as temperature and humidity. Therefore, we conduct online estimation of the UWB systematic range errors. The UWB measurement of the $i$-th base station can be modeled as

$$\hat{d}_i = s_{u_i} d_i + b_{u_i} + n_i, \quad (10)$$

where $\hat{d}_i$ and $d_i$ is the UWB range measurement and the true distance from the tag to the $i$-th base station, respectively; $s_{u_i}$ and $b_{u_i}$ are the scale factor and bias, respectively; the noise $n_i$ is a zero mean Gaussian random process. $s_{u_i}$ and $b_{u_i}$ are modeled as random walks and the corresponding error states are augmented into the state vector as (1) and (2). Hence, the UWB systematic range errors are estimated online with the assistance of IMU and LiDAR. The estimated systematic range errors will be used to correct range measurements.

For the range measurement $\hat{d}_{i,j}$ of the $i$-th base station at the $j$-th UWB keyframe time, the residual can be written as

$$r_{i,j}^{UWB}(\boldsymbol{p}_{u_i}^w, \delta \boldsymbol{x}) = \hat{d}_{i,j} - (\hat{s}_{u_i} \| \hat{\boldsymbol{p}}_{b_j}^w + \hat{\mathbf{R}}_{b_j}^w \boldsymbol{p}_{tag}^b - \boldsymbol{p}_{u_i}^w \| + \hat{b}_{u_i}), \quad (11)$$

where $\{\hat{\mathbf{R}}_{b_j}^w, \hat{\boldsymbol{p}}_{b_j}^w\}$ is the IMU pose at the $j$-th UWB keyframe time; $\boldsymbol{p}_{tag}^b$ is the pre-calibrated extrinsic parameter between the IMU and UWB tag; $\boldsymbol{p}_{u_i}^w$ is the position of the $i$-th base station in the $w$ frame. Then, we can derive the corresponding analytical Jacobians w.r.t $\{\delta s_{u_i}, \delta b_{u_i}\}$ and $\{\delta \boldsymbol{\theta}_{u_j}^w, \delta \boldsymbol{p}_{u_j}^w\}$. Finally, the UWB measurement update is conducted with the standard ESKF update formula [25].

### D. State Augmentation and Marginalization

Compared with the standard ESKF with fixed states, the adopted MSCKF requires extra state augmentation and marginalization. After the LiDAR/ UWB measurement update, the LiDAR/ UWB pose state is augmented into the state vector, and the corresponding covariance $\mathbf{P}_{n \times n}$ is augmented as

$$\mathbf{P}_{(n+6) \times (n+6)} = \begin{bmatrix} \mathbf{I}_{n \times n} \\ \mathbf{J}_{6 \times n} \end{bmatrix} \mathbf{P}_{n \times n} \begin{bmatrix} \mathbf{I}_{n \times n} \\ \mathbf{J}_{6 \times n} \end{bmatrix}^T, \quad (12)$$

where $\mathbf{J}_{6 \times n}$ is the Jacobian of the augmented pose state w.r.t the state vector. The state and covariance of the oldest LiDAR/ UWB keyframe will be directly deleted when the LiDAR/ UWB sliding window exceeds its maximum length, i.e. the marginalization [28].

## IV. RANSAC-BASED MULTI-EPOCH OUTLIER REJECTION FOR UWB RANGES

UWB ranges are severely affected by environmental factors such as NLOS and multipath in complex environments, which can lead to outliers, thereby affecting positioning accuracy. Therefore, we need to remove outliers from UWB ranges. In the short term, the relative-pose accuracy of the LiDAR-inertial odometry is high, and the short-term trajectory can be considered consistent with the true trajectory. Therefore, the consistency of multi-epoch range measurements in the UWB sliding window can be checked using the relative trajectory of the LIO.

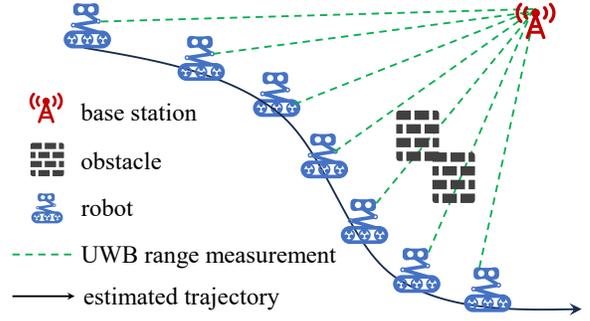

Fig. 3. UWB multi-epoch measurements.

### A. UWB Base-Station Estimation Model

For the $i$-th UWB base station, there are $Q$ range measurements in the UWB sliding window as shown in Fig. 3, denoted as $D_i = \{d_{i,1}, d_{i,2}, \cdots, d_{i,Q}\}$, where $d_{i,j}$ represents the corrected range measurement for the systematic range errors of the $i$-th base station at the $j$-th UWB keyframe time. The poses at the range measurement time can be represented as $\mathrm{T} = \{(\mathbf{R}_{b_1}^w, \boldsymbol{p}_{b_1}^w), (\mathbf{R}_{b_2}^w, \boldsymbol{p}_{b_2}^w), \cdots, (\mathbf{R}_{b_Q}^w, \boldsymbol{p}_{b_Q}^w)\}$. The position of the $i$-th base station can be calculated using the range from the tag to the base station and the tag pose. The nonlinear optimization method is employed to solve the base station position. The objective function is

$$F(\boldsymbol{p}_{u_i}^w) = \min_{\boldsymbol{p}_{u_i}^w} \sum_{j=1}^{Q} (\| \boldsymbol{p}_{b_j}^w + \mathbf{R}_{b_j}^w \boldsymbol{p}_{tag}^b - \boldsymbol{p}_{u_i}^w \| - d_{i,j})^2. \quad (13)$$

The Levenberg-Marquardt algorithm (LM) is used for solving, and the steps are as follows:

(1) Initialize the initial position $\boldsymbol{p}_{u_i}^{w,(0)}$, the iteration number $k=0$, the maximum number of iterations $K$, the threshold $\epsilon$, and the damping factor $\lambda^{(0)}$.

(2) Calculate the residuals and Jacobians. At the current position $\boldsymbol{p}_{u_i}^{w,(k)}$, the residual $r_j^{(k)}$ and Jacobian $\mathbf{J}_j^{(k)}$ of $d_{i,j}$ can be written as

$$r_j^{(k)} = \| \boldsymbol{p}_{b_j}^w + \mathbf{R}_{b_j}^w \boldsymbol{p}_{tag}^b - \boldsymbol{p}_{u_i}^{w,(k)} \| - d_{i,j} \quad (14)$$

$$\mathbf{J}_j^{(k)} = \frac{\partial r_j^{(k)}}{\partial \boldsymbol{p}_{u_i}^w} = -\frac{(\boldsymbol{p}_{b_j}^w + \mathbf{R}_{b_j}^w \boldsymbol{p}_{tag}^b - \boldsymbol{p}_{u_i}^{w,(k)})^T}{\| \boldsymbol{p}_{b_j}^w + \mathbf{R}_{b_j}^w \boldsymbol{p}_{tag}^b - \boldsymbol{p}_{u_i}^{w,(k)} \|} \quad (15)$$

Hence, the residual $\boldsymbol{r}^{(k)}$ and Jacobian matrix $\mathbf{J}^{(k)}$ of the objective function can be written as

$$\boldsymbol{r}^{(k)} = [r_1^{(k)}, r_2^{(k)}, \cdots, r_Q^{(k)}]^T, \mathbf{J}^{(k)} = [\mathbf{J}_1^{(k)T}, \mathbf{J}_2^{(k)T}, \cdots, \mathbf{J}_Q^{(k)T}]^T. \quad (16)$$

(3) Solve the following linear equation to solve the increment $\Delta \boldsymbol{p}_{u_i}^{w,(k)}$ by QR or Cholesky decomposition:

$$(\mathbf{J}^{(k)T} \mathbf{J}^{(k)} + \lambda^{(k)} \mathbf{I}) \Delta \boldsymbol{p}_{u_i}^{w,(k)} = -\mathbf{J}^{(k)T} \boldsymbol{r}^{(k)}. \quad (17)$$

(4) Update the base station position as

$$\boldsymbol{p}_{u_i}^{w,(k+1)} = \boldsymbol{p}_{u_i}^{w,(k)} + \Delta \boldsymbol{p}_{u_i}^{w,(k)}. \quad (18)$$

**Algorithm 1:** RANSAC-based outlier rejection

**Input**: UWB range dataset $D_i$, base-station estimation model $F$, minimum number $P$ of samples, maximum number $K$ of iterations, threshold $\epsilon$, minimum number $L$ of measurements that conform to the model.
**Output:** Optimal model $F_{\text{optimal}}$ and dataset $D_{\text{optimal}}$.

1  $k \leftarrow 0$, $F_{\text{optimal}} \leftarrow \varnothing$, $D_{\text{optimal}} \leftarrow \varnothing$, $\epsilon_{\text{optimal}} \leftarrow \infty$
2  while $k < K$
3      $D_{\text{maybe}} \leftarrow$ randomly select $P$ samples from $D_i$
4      $F_{\text{maybe}} \leftarrow$ fit model $F$ with $D_{\text{maybe}}$
5      $D_{\text{also}} \leftarrow \varnothing$
6      for each $d \in D_i$ and $d \notin D_{\text{maybe}}$
7         if the error of $d$ for $F_{\text{maybe}}$ is less than $\epsilon$
8            $D_{\text{also}} \leftarrow D_{\text{also}} \cup \{d\}$
9      if $|D_{\text{also}}| > L$
10        $F_{\text{better}} \leftarrow$ fit model $F$ with $D_{\text{maybe}} \cup D_{\text{also}}$
11        $\epsilon_{\text{better}} \leftarrow$ the sum of errors for $F_{\text{better}}$
12        if $\epsilon_{\text{better}} < \epsilon_{\text{optimal}}$
13           $F_{\text{optimal}} \leftarrow F_{\text{better}}$, $\epsilon_{\text{optimal}} \leftarrow \epsilon_{\text{better}}$
14           $D_{\text{optimal}} \leftarrow D_{\text{maybe}} \cup D_{\text{also}}$
15     $k \leftarrow k+1$
16 return $F_{\text{optimal}}, D_{\text{optimal}}$

(5) Check the termination condition. If the increment norm $\|\Delta \boldsymbol{p}_{u_i}^{w,(k)}\|$ is less than the threshold $\epsilon$, or the iteration number reaches the maximum $K$, stop the iteration; otherwise, update $k = k+1$ and return to step 2.

*B. RANSAC-Based Multi-Epoch Outlier Rejection*

The position of the base station relative to the tag can be estimated using the above steps. However, there are outliers in the UWB range measurements resulting from environmental factors such as NLOS and multipath, which may affect the estimation of the base station position. Therefore, we employ the RANSAC algorithm to remove the UWB range outliers as shown in Algorithm 1. Specifically, $P$ ranges are randomly sampled from $D_i$ and used for base-station position estimation with the above model. The residuals of other ranges in $D_i$ are calculated using the estimated base station position. Then, the ranges with residuals less than $\epsilon$ and the sampled ranges are used for a new round of base-station position estimation, and the sum of their residuals is calculated with the new estimated base-station position. After $K$ iterations, the base-station position estimation with the smallest residual sum is selected as the optimal model, and the ranges used are taken as the optimal dataset $D_{\text{optimal}}$. Finally, the UWB ranges in $D_{\text{optimal}}$ are used for UWB measurement update.

## V. EXPERIMENTS AND RESULTS

*A. Experimental Setups*

A wheeled robot with a maximum speed of 1.5 m/s is used

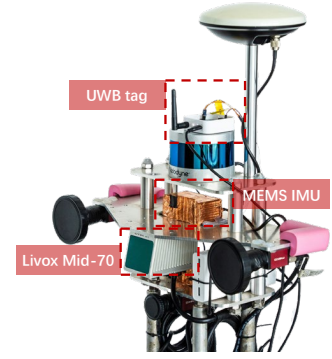

Fig. 4. The experimental platform.

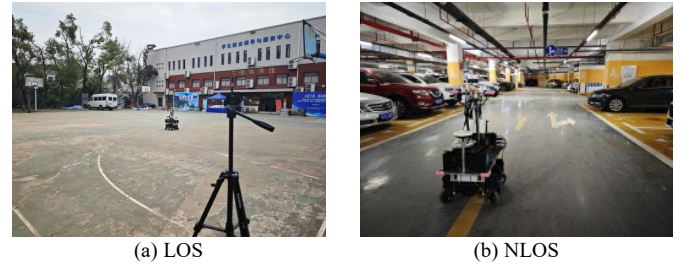

(a) LOS  (b) NLOS
Fig. 5. Testing environments.

for data collection. The main experimental equipment includes a MEMS IMU ADIS16465, a solid-state LiDAR Livox Mid-70, a UWB tag, and four UWB base stations, as shown in Fig. 4. The lever-arm between the UWB tag and IMU is precisely measured. The UWB tag, IMU, and LiDAR are well-synchronized. The UWB device adopts the DS-TWR [27] method for ranging, with a standard deviation of 3 cm and a frequency of 5 Hz.

We conducted real-world experiments in both UWB LOS and NLOS environments. The LOS testing environment is an outdoor open field, as shown in Fig. 5 (a), which is used to test the UWB positioning accuracy in the ideal environment; the NLOS testing environment is an underground garage with serious UWB NLOS interference, as shown in Fig. 5 (b). The base station distributions and testing trajectories are shown in Fig. 6. In the LOS and NLOS environments, spatial distributions of UWB base stations are the same to eliminate the impact of base station distribution on UWB positioning accuracy. In both environments, four UWB base stations are distributed at the four corners of a 22×24 m rectangle, with a distance of 24 m between A0 and A1, and a distance of 22 m between A0 and A3. In the LOS tests, there is no obstruction; in the NLOS tests, obstacles such as vehicles and walls will cause UWB NLOS signals as the UWB base stations are mounted on low-height tripods.

Three tests are conducted in each environment, and the testing trajectories are represented by different colors in Fig. 6. The range residuals calculated with the ground truth and the number of UWB ranges in the *NLOS-01* sequence are shown in Fig. 7. The tag could not observe four base stations at the same time most of the time, and the range measurements were severely affected by NLOS, resulting in significant residuals. It should be noted that there is a LiDAR-degenerated scene in the *NLOS-02* sequence, *i.e.*, a flat wall which is shown by the vertical line on the left of Fig. 6 (b). When the robot turns at this location, the LiDAR can only observe the wall and ground due to its limited field of view, resulting in insufficient constraints in the horizontal direction. In addition, we simulated the case where only two base stations are available in the NLOS environment to evaluate the performance of MR-ULINS under conditions with sparse UWB base stations.

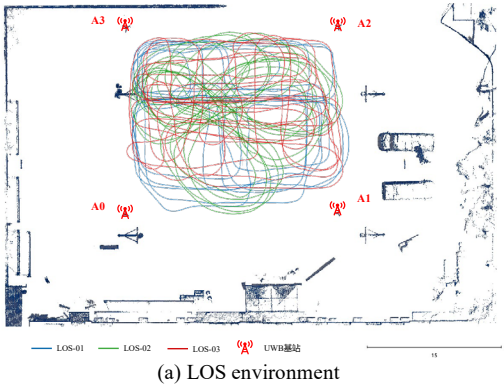
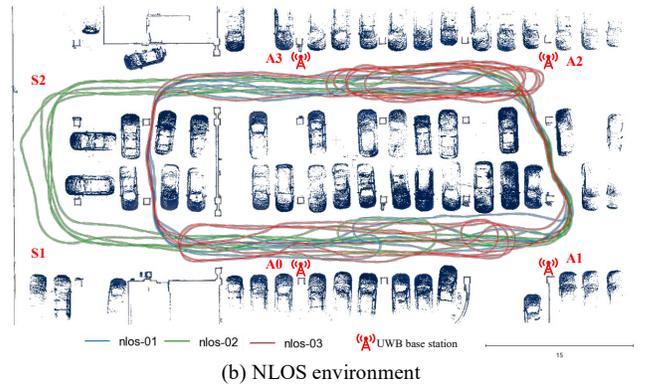

(a) LOS environment                (b) NLOS environment

Fig. 6. Base station distributions and testing trajectories. The distance between *A0* and *A1*, *A2* and *A3* is about 24 m, and the distance between *A0* and *A3*, *A1* and *A2* is 22 m. *S1* and *S2* are where LiDAR degradation occurs.

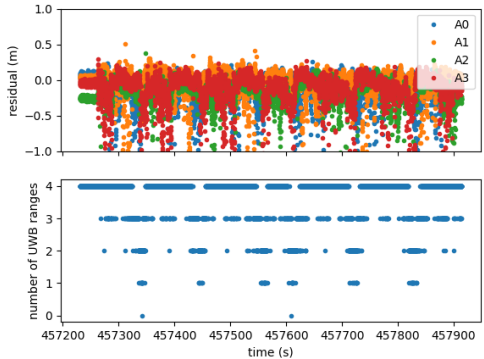

Fig. 7. UWB range residuals and numbers of the *NLOS-01* sequence.

TABLE I
RMSEs IN THE LOS ENVIRONMENT

| Sequence | FAST-LIO2 | TC-LIO | UINS | ULINS | MR-ULINS |
|---|---|---|---|---|---|
| *LOS-01* | 0.365 | 0.371 | 0.098 | **0.090** | **0.090** |
| *LOS-02* | 0.225 | 0.839 | 0.117 | **0.104** | 0.109 |
| *LOS-03* | 0.378 | 0.688 | 0.120 | 0.094 | **0.093** |
| RMS | 0.330 | 0.662 | 0.112 | **0.096** | 0.098 |

TABLE II
RMSEs IN THE NLOS ENVIRONMENT

| Sequence | FAST-LIO2 | TC-LIO | ULINS | MR-ULINS |
|---|---|---|---|---|
| *NLOS-01* | 0.204 | 0.557 | 0.157 | **0.088** |
| *NLOS-02* | 0.290 | 0.659 | 0.196 | **0.122** |
| *NLOS-03* | 0.215 | 0.313 | 0.145 | **0.097** |
| RMS | 0.239 | 0.530 | 0.167 | **0.103** |

In the LOS experiments, the integrated navigation solution of a navigation-grade [24] IMU and GNSS real-time kinematic (RTK) is used as the ground truth; in the NLOS experiments, the ground truth comes from a laser mapping device RS100i-MT designed by GoSLAM, whose mapping accuracy is 1 cm. The point clouds collected by Livox Mid-70 are matched with the point-cloud map constructed by RS100i-MT to solve the robot poses as the ground truth. We evaluate the positioning accuracy by comparing the estimated trajectories with the ground truths. The metrics include root mean square error (RMSE) and cumulative distribution function (CDF) of the positioning error.

We implement the proposed MR-ULINS in C++ and Robots Operating System (ROS). The algorithms used for comparison include 1) FAST-LIO2 [29]: one of the state-of-the-art (SOTA) LIOs, which employs frame-to-map (F2M) data association; 2) UINS: tightly-coupled UWB-inertial system; 3) TC-LIO: tightly-coupled LiDAR-inertial odometry with F2F data association; 4) ULINS: tightly-coupled UWB-LiDAR-inertial system, without the proposed UWB systematic range-error estimation or multi-epoch outlier rejection, with only single-epoch outlier rejection, *i.e.*, the chi-squared test.

All the algorithms are implemented on a desktop PC (Intel i9-11900K).

*B. Accuracy Evaluation*

*1) LOS Environment*

The RMSEs in the LOS environment are shown in TABLE I. The algorithms using UWB exhibit higher accuracy than LIOs (FAST-LIO2 and TC-LIO) because UWB can provide accurate and drift-free absolute positioning in the LOS environment. The root mean square (RMS) of MR-ULINS RMSEs is 0.098 m and smaller than UINS and the baseline ULINS, indicating that the proposed MR-ULINS has better positioning accuracy in the ideal environment for UWB.

*2) NLOS Environment*

The RMSEs in the NLOS environment are shown in TABLE II, and the positioning-error CDF curves of the *NLOS-02* sequence are shown in Fig. 8. The results of UINS are not shown since UINS exhibits poor accuracy in the NLOS environment. Due to many loop closures in the NLOS sequences, FAST-LIO2 based on F2M association can establish associations with the self-built map, resulting in high accuracy. TC-LIO exhibits lower accuracy than FAST-LIO2, especially in the *NLOS-02* sequence. The reason is that the robot repeatedly turned at the flat wall on the left of Fig. 6 (b). When the robot turned at this location, the F2F association method employed by TC-LIO could not provide sufficient constraints in the horizontal direction due to the limitation of the Livox Mid-70 field of view. Compared with FAST-LIO2 and TC-LIO, ULINS and MR-ULINS mitigate the cumulative errors using the absolute range measurements of UWB. MR-ULINS reduces the systematic range errors and NLOS of UWB, thus exhibiting the best performance in the NLOS environment according to the RMSEs and CDFs. Compared with the baseline ULINS, the average RMSE of MR-ULINS is reduced by 38.3%.

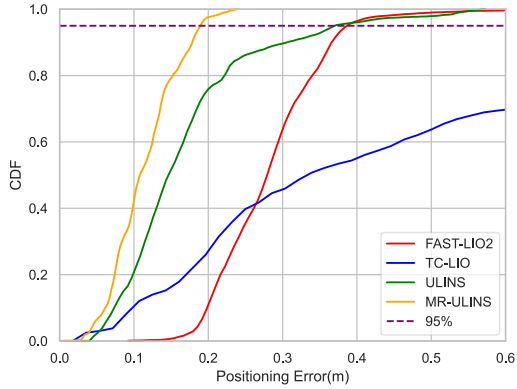

Fig. 8. Position error CDFs for the *NLOS-02* sequence.

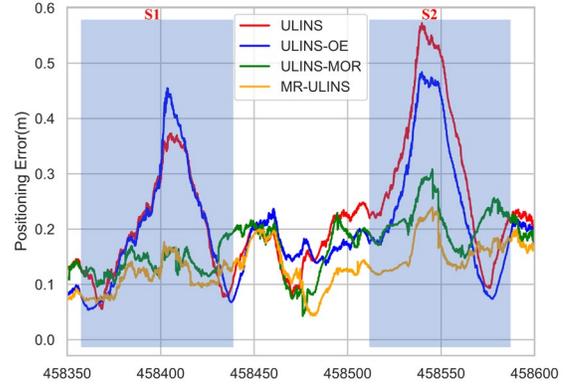

Fig. 9. Positioning error curves for the LiDAR-degenerated scenes *S1* and *S2* in the NLOS-02 sequence.

TABLE III
RMSEs OF THE ABLATION EXPERIMENTS

| Sequence | ULINS | ULINS-OE | ULINS-MOR | MR-ULINS |
|---|---|---|---|---|
| *NLOS-01* | 0.157 | 0.115 | 0.130 | **0.088** |
| *NLOS-02* | 0.196 | 0.163 | 0.150 | **0.122** |
| *NLOS-03* | 0.145 | 0.128 | 0.127 | **0.097** |
| RMS | 0.167 | 0.137 | 0.136 | **0.103** |

TABLE IV
RMSEs IN SPARSE UWB BASE-STATION CONDITIONS

| sequence | ULINS | ULINS-OE | ULINS-MOR | MR-ULINS |
|---|---|---|---|---|
| Test 1: only use base stations *A1* and *A3* | | | | |
| *NLOS-01* | 0.224 | 0.132 | 0.112 | **0.107** |
| *NLOS-02* | 0.397 | 0.454 | 0.145 | **0.143** |
| *NLOS-03* | 0.177 | 0.154 | 0.127 | **0.102** |
| RMS | 0.282 | 0.287 | 0.129 | **0.119** |
| Test 2: only use base stations *A0* and *A2* | | | | |
| *NLOS-01* | 0.424 | 0.170 | 0.175 | **0.097** |
| *NLOS-02* | 0.678 | 0.225 | 0.187 | **0.123** |
| *NLOS-03* | 0.286 | 0.185 | 0.162 | **0.103** |
| RMS | 0.490 | 0.195 | 0.175 | **0.108** |
| Test 3: only use base stations *A0* and *A3* | | | | |
| *NLOS-01* | 0.231 | 0.141 | 0.146 | **0.099** |
| *NLOS-02* | 0.333 | 0.302 | 0.155 | **0.122** |
| *NLOS-03* | 0.169 | 0.127 | 0.142 | **0.111** |
| RMS | 0.254 | 0.206 | 0.148 | **0.111** |

The accuracy of ULINS is affected by UWB systematic range errors and NLOS, resulting in a significant increase in the positioning errors compared with the LOS experiment results (from 0.096 m to 0.167 m). In contrast, the accuracy of MR-ULINS in the NLOS environment is comparable to that in the ideal LOS environment since the online estimation of range errors and multi-epoch outlier rejection improve the accuracy of UWB range measurements significantly.

*C. Ablation Experiments*

The proposed MR-ULINS adds online estimation of UWB systematic range errors and multi-epoch outlier rejection on the baseline ULINS. Hence, we conducted ablation experiments to verify their respective impacts on the positioning accuracy. ULINS with online estimation of systematic range errors is denoted as ULINS-OE, and ULINS with multi-epoch outlier rejection is denoted as ULINS-MOR. The results of the ablation experiments are shown in TABLE III. Compared with the baseline ULINS, ULINS-OE and ULINS-MOR both exhibit higher accuracy, with average RMSE reduction rates of 18.0% and 18.6%, respectively, which indicates that both algorithms improve the positioning accuracy. Besides, the proposed MR-ULINS combines the online estimation and multi-epoch outlier rejection, and achieves the optimal positioning accuracy, indicating that the improvement effects of the two algorithms are complementary.

In the *NLOS-02* sequence, the positioning error curves during the LiDAR-degenerated scenes *S1* and *S2* in Fig. 6 (b) are shown in Fig. 9. The positioning errors of ULINS-MOR are significantly lower than ULINS and ULINS-OE which use the single-epoch outlier rejection, indicating that the multi-epoch outlier rejection improves the positioning accuracy significantly in the LiDAR-degenerated scene. When the LiDAR degradation occurs, the estimated absolute pose becomes inaccurate. The single-epoch outlier rejection may mistake normal measurements for abnormal measurements. Hence, the cumulative errors caused by LiDAR degradation cannot be corrected in time. In contrast, multi-epoch outlier rejection utilizes the short-term high-precision relative trajectory of the LIO to check the consistency of multi-epoch UWB ranges. Thus, it can effectively remove range outliers even if the absolute pose is inaccurate.

*D. Robustness Evaluation with Sparse UWB Base Station*

The tightly coupled MR-ULINS can achieve drift-free positioning with fewer UWB base stations than four. Hence, we further conduct sparse base-station experiments in the NLOS environment to evaluate the robustness of MR-ULINS. Three experiments were conducted, only using the range measurements of base stations *A1* and *A3*, *A0* and *A2*, and *A0* and *A3*, respectively. The results are shown in TABLE IV. Under conditions with sparse base stations, the improvement of online estimation and multi-epoch outlier rejection on the positioning accuracy is more significant. Combining the two algorithms, MR-ULINS achieves superior accuracy than the baseline ULINS and exhibits higher accuracy than ULINS-OE and ULINS-MOR in most cases. The RMSEs of MR-ULINS still maintain at around 0.1 m, comparable to the errors when using four base stations. Hence, MR-ULINS can reduce the base-station density and deployment costs without sacrificing the positioning accuracy.

TABLE V
AVERAGE RUNNING TIMES OF MR-ULINS ON THE NLOS SEQUENCES

| Time (ms) | NLOS-01 | NLOS-01 | NLOS-01 |
|---|---|---|---|
| forward propagation | 11.4 | 11.7 | 11.0 |
| LiDAR update | 29.6 | 30.0 | 29.3 |
| outlier rejection | 21.4 | 21.4 | 21.1 |
| UWB update | 3.1 | 3.2 | 3.1 |

*E. Efficiency Evaluation*

The average running times of MR-ULINS on the NLOS sequences are shown in TABLE V. The average intervals of the LiDAR keyframes and UWB keyframes are around 300 ms and 600 ms, respectively. Hence, the results in TABLE V indicate that MR-ULINS exhibits superior real-time performance. According to the statistical results, MR-ULINS can run at around 5 times the real-time speed when the sliding window sizes of LiDAR and UWB are both set to 20.

VI. CONCLUSION

This paper proposes MR-ULINS, a tightly-coupled UWB-LiDAR-inertial estimator within the MSCKF framework. The LiDAR measurement model constructs a relative constraint by F2F data association. The absolute positioning sensor UWB can be seamlessly integrated to eliminate cumulative errors. The UWB systematic range errors are precisely modeled to be estimated and calibrated online and thus the positioning accuracy is improved. Meanwhile, the impact of the UWB NLOS is reduced notably by employing the proposed multi-epoch outlier rejection. MR-ULINS exhibits similar accuracy, *i.e.* about 0.1 m, in both the LOS and NLOS environments with the same distribution conditions of UWB base stations. Besides, MR-ULINS demonstrates superior robustness in LiDAR-degenerated scenes and conditions with sparse UWB base stations. We will integrate the GNSS to achieve seamless indoor-outdoor positioning in future works.